**Deep Lexical Hypothesis:**

**Identifying personality structure in natural language**

Andrew Cutler[1] and David M. Condon[2]

[1] Department of Electrical Engineering, Boston University

[2] Department of Psychology, University of Oregon

```
Draft dated February 21, 2022.
Note: this version has not been peer reviewed, and
 may not be the authoritative document of record.
```

**Author Note**

Andrew Cutler  https://orcid.org/0000-0002-6616-8522
David M. Condon  https://orcid.org/0000-0002-8406-783X

Correspondence concerning this article should be addressed to David M. Condon, Department of Psychology, University of Oregon, 1451 Onyx Street, Eugene, Oregon, USA, 97403. Email: dcondon@uoregon.edu

**Abstract**

Recent advances in natural language processing (NLP) have produced general models that can perform complex tasks such as summarizing long passages and translating across languages. Here, we introduce a method to extract adjective similarities from language models as done with survey-based ratings in traditional psycholexical studies but using millions of times more text in a natural setting. The correlational structure produced through this method is highly similar to that of self- and other-ratings of 435 terms reported by Saucier and Goldberg (1996a). The first three unrotated factors produced using NLP are congruent with those in survey data, with coefficients of 0.89, 0.79, and 0.79. This structure is robust to many modeling decisions: adjective set, including those with 1,710 terms (Goldberg, 1982) and 18,000 terms (Allport & Odbert, 1936); the query used to extract correlations; and language model. Notably, Neuroticism and Openness are only weakly and inconsistently recovered. This is a new source of signal that is closer to the original (semantic) vision of the Lexical Hypothesis. The method can be applied where surveys cannot: in dozens of languages simultaneously, with tens of thousands of items, on historical text, and at extremely large scale for little cost. The code is made public to facilitate reproduction and fast iteration in new directions of research.

**Deep Lexical Hypothesis: Identifying personality structure in natural language**

Understanding the comprehensive structure of psychological individual differences offers the potential to develop models that can describe, predict, and explain the ways that important life outcomes are (and are not) shaped by personality (Mõttus et al., 2020). Indeed, the utility of models of between-person differences is substantial, as evidenced by the influence of the Big Few models of personality over the last 30 years (Ozer & Benet-Martinez, 2006; Roberts et al., 2007; Soto, 2019). These models include, most prominently, the Big Five (Goldberg, 1990), the Five-Factor Model (McCrae & Costa, 1987), the HEXACO (Ashton, Lee, Perugini, et al., 2004), and the Big Six (Thalmayer et al., 2011). The personality literature is replete with research findings about the extent to which the broad-bandwidth variables that comprise these models affect many domains of life, including features of self-concept and identity; one's relationships with peers, friends, and romantic partners; mental health, including well-being and the experience of psychopathology; factors affecting physical health and longevity; choices relating to occupational and educational trajectories (as well as performance in those domains); and the extent and manner of engagement with broader social issues and institutions (Ozer & Benet-Martinez, 2006; Soto, 2019). The application of these models has been so extensive that it may even prompt questions regarding the need for further research on personality structure, especially at the highest levels of parsimony (i.e., just a few dimensions).

Yet, several recent contributions have highlighted unresolved issues. The most notable concerns relate to generalizability, especially in terms of the replicability and universal relevance of the Big Few traits (Allik & Realo, 2017; Cheung, van de Vijver, & Leong, 2011; De Raad & Peabody, 2005; De Raad et al., 2010; Gurven et al., 2013; McCrae, Terracciano, & Personality

Profiles of Cultures Project, 2005a, 2005b). Though it has been argued that generalizability follows from early evidence of similar patterns of covariation among the Big Few traits across cultures (McCrae, 2009; McCrae & Costa, 1997), others have noted more recently that consistent covariation can mask differences in salience (i.e., differences in means) or the exclusion of meaningful content within these broad factors (Allik & Realo, 2017; Cheung et al., 2011; Saucier et al., 2014). Further, several recent efforts to replicate Big Few structures using exploratory approaches have failed, particularly in contexts outside of Western cultures (Daouk-Öyry et al., 2016; De Raad et al., 2014; Gurven et al., 2013; Laajaj et al, 2019; Thalmayer, Saucier, Ole-Kotikash, et al., 2020; Wood et al., 2020). Many cultures have never been robustly evaluated (Thalmayer, Toscanelli, & Arnett, 2020). Pragmatists on both sides of this issue acknowledge the shortcomings of the Big Few models while also pointing out that they are preferable to the difficulties caused by the chaotic lack of consensus that preceded the advent of the Big Five (Goldberg, 1993; Mõttus et al., 2017). Broad adoption of these models has allowed the field to advance the goal of demonstrating the robust utility of personality trait-outcome research (Atherton et al., 2021; Soto, 2019). Still, continued emphasis on this priority has the potential to discourage further advancements beyond the Big Few in the absence of "empirical comparison of alternative scientific models" (p. 221, Goldberg & Saucier, 1995). In other words, the identification of a more generalizable model grows more challenging over time as consensus around the Big Few is further instantiated by personality-outcome research that nearly always fails to compare validity across models. The task of making ever-more compelling claims about the utility of personality traits may require an interim period of increased divergence in measurement, as new models are identified, developed, and tested.

A more nuanced concern, related to generalizability, stems from the nature of the content being assessed and/or left out by existing trait models. Several researchers have criticized the Big Few on the grounds that the small number of dimensions requires them to be overly broad (Block, 1995a; Condon et al., 2021; Ones & Viswesvaran, 1996). The benefits of parsimony – specifically, the ability to describe the rich complexity of psychological individual differences with only five or six constructs – comes at the expense of being required to use highly abstracted dimensions. Several research teams have sought to address this problem by developing measures of more narrow constructs that are hierarchically nested within the Big Few dimensions (Ashton, Lee, Perugini, et al., 2004; Costa & McCrae, 1995; DeYoung, Quilty, & Peterson, 2007; Soto & John, 2017). The resulting assessment models contain varying numbers of constructs (10 to 15 aspects or 24 to 30 facets). Unfortunately, hierarchically nested approaches do not fully address the issue of omitted content because they are derivative of Big Few models at the highest level (Condon, 2018; Saucier & Iurino, 2019). They help, to some extent, by covering the same scope of content with more narrowly focused measures, but they do not account for constructs that fall outside the Big Few dimensions (e.g., humor, aggression; Paunonen & Jackson, 2000), including those from other domains of individual differences research (e.g., socio-emotional skills, interests, values, cognitive abilities; Ackerman, 2018; Ackerman & Heggestad, 1997; Condon, 2014; Soto, Napolitano, Sewell, Yoon, & Roberts, 2021). Their derivative nature means they also inherit the aforementioned concerns about generalizability. Resolution of this issue would seem to require the development of a comprehensive and well-specified taxonomy of traits (Allport & Odbert, 1936; Condon et al., 2021; Goldberg, 1982) – an aspirational goal for decades.

Collectively, these concerns provide strong motivation for ongoing research on personality structure generally, though it is unclear whether the concerns will be resolved by further rounds of data collection using survey-based ratings of English adjectives. In addition to being time-consuming and costly, traditional techniques in psycholexical research rely on methods that constrain their generalizability. As discussed in more detail below, these include limitations in the sampling of variables and populations, and attributes that make these methods poorly suited to the study of multiple languages simultaneously (Wood et al., 2020).

The current work departs from the approach used over the last several decades by drawing on recent innovations in natural language processing (NLP). Since the introduction of "transformer" language models like Google's BERT (Devlin et al., 2019), it has become possible to conduct powerful evaluations of the associations between words and other aspects of language in ways that are novel and likely to be insightful for psychological science (Jackson et al., 2021). The data underlying these models is different from traditional personality data in that it comes from multiple large corpora of text rather than adjective ratings, but both types are intended to reflect the structure of personality descriptors. As language modeling technologies also have the potential to alleviate many of the difficulties stemming from survey-based data collection on personality descriptors, an important goal is to rigorously evaluate whether – and how much – personality structure is consistent across the two data sources. Before addressing this research question empirically, we begin by describing the potential benefits of using language model output (an "NLP approach") relative to the use of survey-based self- and other-report ratings.

**Traditional and NLP approaches to psycholexical research**

The long history of psycholexical research on the structure of personality rests on the assumption that personality traits are instantiated by person-descriptive terms in the lexicon – trait descriptive adjectives and "type" nouns (Allport & Odbert, 1936; Galton, 1884; Goldberg,

1993; McCrae, 1990; Saucier & Goldberg, 1996b).[1] Language allows humans to describe the specific ways in which we differ (e.g., "talkative", "witty", "rude", or "a loner"). On the basis of this idea, often referenced as the lexical hypothesis (or postulate), personality psychologists have made numerous attempts to identify a parsimonious and replicable description of the multidimensional space of personality traits (Allport & Odbert, 1936; Ashton, Lee, & Goldberg,

2004; Cattell, 1943; Digman & Inouye, 1986; Goldberg, 1981, 1990, 1992; McCrae, 1990;

McCrae & Costa, 1985; Norman, 1963; Thurstone, 1934; Tupes & Christal, 1961). Extended

descriptions of the procedures used to carry out these efforts have previously been provided

(Angleitner et al., 1990; De Raad & Hendriks, 1997; Goldberg, 2006; Goldberg & Digman,

1994; Saucier, 1997; Saucier & Goldberg, 2001), but they can be summarized as having three

essential steps: (i) subsetting the person-descriptors to be analyzed; (ii) collecting data on the

subset of terms from a large and representative sample of the targeted population; and (iii)

reducing the (often unwieldy) dimensionality of the resulting data into a tractable model. In the following subsections, we review the ways that these steps have been completed traditionally, and how they may be altered by the use of NLP.

*Step 1: Subsetting the universe of descriptors*

         The first stage involves identification of some subset of personality descriptors for use as

stimuli in data collection with large samples (Cattell, 1957; Norman, 1963; Peabody & Goldberg,

1989). Subsetting has been required because the complete universe of personality descriptors includes many thousands of terms (at least 17,953; Allport & Odbert, 1936), and it is infeasible to administer more than a few hundred in the same sample (barring the use of planned missingness procedures that affect sample size considerations; Revelle et al., 2016). The largest subset reported to date (Ashton, Lee, and Goldberg, 2004; Saucier & Iurino, 2019) included 1,710 terms identified by Goldberg (1982) and administered to approximately 300 undergraduates from the US (by Goldberg, N=187) and Australia (by Norman, N=123). Most other subsets have been considerably smaller, including those used for development and replication of the Big Five factor structure (587 adjectives, Goldberg, 1990; 435 adjectives, Saucier & Goldberg, 1996a), and development of the HEXACO measure (449 adjectives; Lee & Ashton, 2004).

---

[1] The vast majority of traditional psycholexical work has focused on adjectives and the relatively small
number of type-nouns, though personality relevant information is also conveyed in adverbs and verbs
(and possibly other parts of speech). For example, see Angleitner and Riemann (1991).

The primary means of identifying subsets has been qualitative judgments about the properties of words and the relations between them (Cattell, 1945). These procedures have typically been conducted by one- or two-person research teams working independently, removing terms on the basis of synonymity (De Raad & Mlačić, 2017), perceived ambiguity, irrelevance to personality, or lack of familiarity (Ashton & Lee, 2004; Condon et al., 2022), or similar criteria (e.g., evaluativeness; Goldberg, 1982). This qualitative approach has been criticized for introducing subjectivity and bias (Block, 1995; Digman & Takemoto-Chock, 1981; Horn, 1967; Tellegen, 1993). These criticisms are supported by evidence that decisions about subsetting can meaningfully alter subsequent findings about structure (Ashton, Lee, & Goldberg, 2004; Saucier, 1994). Though more formal methods are possible, such as clustering techniques for synonymity (Revelle, 1979), these require data collection that obviates the need to subset.

NLP approaches circumvent the need to subset. This is possible because NLP models represent words as vectors of several hundred to several thousand dimensions (i.e., word embeddings; see "word2vec"; Mikolov et al., 2013) such that the meaning of each word is encoded based on features of the source text. In traditional methods, words are vectorized with ratings – each person provides a value for each term – and this limits the number of terms under study. NLP models can be used to generate similar data structures. While traditional methods yield matrices in which each row represents a person, the rows of NLP models, by contrast, are features or sets of features identified during the training phase of an algorithm. However, the same underlying logic applies to both approaches: trait descriptors with similar weightings across features (or ratings across people) will be highly correlated (discussed further below).

When using previously developed (i.e., "pre-trained") language models, it is possible to immediately query the relationships among word sets of virtually any size. Analyses based on larger sets (i.e., 18,000 terms or more) are only slightly more time-consuming than those based on the 300 to 500 word sets used historically in personality. This provides at least two benefits. First, it means that exploratory analyses of structure can cover the full scope of descriptors evaluated in the lexicon, allowing for stronger claims to be made about the recovered structure. Second, it provides the ability to conduct sensitivity analyses that are impractical given the costs and time-consuming nature of survey methods. For example, this makes it possible to consider the extent to which structure depends based on the use of many different subsets of descriptors and/or many different language modeling techniques.

An important question – central to the current work – is the extent to which the NLP approach fails to capture personality-relevant information or introduces additional noise to the associations among personality descriptors. To be clear, it is not problematic that language models produce data that reflects semantic structure rather than judgments of behavior, as it is the essence of the lexical hypothesis that these structures are the same. That said, evidence for directional differences in the associations between terms across methods would not necessarily invalidate either approach. Put simply, differences may reflect valid distinctions between psychology and linguistics.

*Step 2: Data collection from a large and representative sample*

      Traditional data collection methods involve asking raters to evaluate how well the terms describe themselves or someone else – friends, family members, strangers, or even others whom they dislike (Connelly & Ones, 2010; McCrae & Costa, 1987; Peabody & Goldberg, 1989; Somer & Goldberg, 1999). The recruitment of raters who are representative of the broader population is particularly important at this stage because the variability among ratings will otherwise be biased by disproportionate homogeneity (Allik & Realo, 2017; Beck et al., 2019). Though increasingly well recognized in recent years (Henrich et al., 2010), the problem of excessive homogeneity in psychological assessment, and psychological science more broadly, is challenging to resolve with survey-based methods. Telemetric data collection reduces many of the limitations present before the internet era (Gosling et al., 2004; Wilt et al., 2012), yet important populations remain hard to reach, including those who lack the resources and awareness needed to participate, as well as those who lack the interest. Further, large-scale representative sampling is more resource intensive than convenience sampling (i.e., undergraduates), introducing further bias in the representativeness of researchers who conduct such studies (Thalmayer, Toscanelli, & Arnett, 2020).

      As a computational technique for analyzing the ideas conveyed in large repositories of written language, the NLP method does not rely on similar data collection procedures. Language models are developed through a series of large-scale, collaborative procedures that begins with the aggregation of large corpora from various sources, text extraction and cleaning, and pre-processing (Gupta et al., 2020; Rae et al., 2021). Though the approaches clearly differ, NLP and

survey-based methods are similar in that the underlying data are contributed by individual people who will influence the resulting structure in proportion to the total informational content of the full sample of contributors. Thurstone (1934) was developing, in essence, a type of language model when instructing respondents to rate themselves using an adjective checklist, as this data generating process is functionally equivalent to the generation of a statement about one's personality (Cutler, 2021). A benefit of Thurstone's approach is that the resulting data are highly structured, but the trade-off is substantial – the pool of respondents is limited to those who can be recruited and motivated to complete the checklist. In contrast, the data used to train modern language models (i.e., corpora) are much less structured, but the constraints on data acquisition can be considerably more relaxed. As a consequence, the data used for many NLP applications (that is, to train the models) are typically larger than the data collected using survey methods by many orders of magnitude. One of the largest models released thus far – the Generative Pre-trained Transformer 3 (GPT-3; Brown et al., 2020) – has been trained on 45 terabytes of text data (Dale, 2021). At one byte per character, this is equivalent to approximately 22.5 trillion 5-letter words. By comparison, the largest data set reported to date for analyses of personality structure in English contained ratings from approximately 300 respondents on 1,710 terms.

Given the size of the corpora used to train these language models, the sources of underlying texts tends to be diverse, though this is an important source of variability. Fortunately, a robust ecosystem of open access or inexpensive corpora and NLP models has developed in recent years, enabling comparisons. The options include resources that have been developed by many of the largest for-profit technology and social media companies (e.g., Facebook, Google, Microsoft) as well as crowd-sourced projects and open access collaborations. For example, the Pile (Gao et al., 2020) – among the largest modern corpora – is an open-source project that aggregates content from 22 smaller, high-quality data sets. These data come from sources such as text crawled from the web since 2008, large catalogs of fiction and non-fiction books, content from approximately 5 million publications on PubMed Central, text from human-generated closed-captioning on YouTube, and many more (for further description, see Gao et al., 2020). But, the Pile is only one example. While some recently developed language models (e.g., some versions of GPT) have been trained on it, most have been trained using data sets with varying degrees of transparency with respect to specifics of the underlying corpora. Often, the source texts are named, but the content is inaccessible due to intellectual property concerns and/or technical constraints (file size, for example). Despite opacity in this regard, most NLP models are available for use in research, and many are publicly accessible for free. In mid-2021, for example, more than 17,000 models were openly accessible through the library maintained by Hugging Face ([huggingface.co](https://huggingface.co)), and these could be filtered by attributes such as the data sets used to train them (545 options), the language(s) included in the underlying data sets (i.e., English, Mandarin, Spanish, etc.; 169 options), and the tasks for which the models were developed (23 options).

To summarize, the differences in data collection methods between the traditional and NLP approaches are substantial. The survey method generates data from a relatively small number of respondents, whereas the NLP approach uses model-generated data that has been trained on corpora containing data from a much larger number of contributors. For researchers, generating data from language models is typically free; high quality survey data is often costly. Importantly, the resulting data is also inequivalent in that much of the information used to train the NLP models is not relevant to personality. It reflects the use of everyday language overall, while survey-based methods capture only self- and other-ratings of personality.

*Step 3: Data reduction analyses*

The third and final stage involves analyses of the relations between the stimuli, identified in Step 2. For surveys this is usually calculating the pairwise correlations between word ratings among respondents. For the NLP approach the distinction between data collection and reduction is fuzzier. Word representations obtained from models are already a highly condensed and nonlinear interpolation over training data. These representations are then correlated. From there, the math is no different. However, it is worth noting that differences in the size, scope, and nature of the data could be relevant to the analyses. For example, the ability to immediately conduct sensitivity analyses without extra cost when using NLP methods may increase researchers' confidence in their findings. Similarly, the utility/efficacy of various analyses may differ across the two approaches based on differences in features of the data.

In sum, the NLP approach can address several of the ways that traditional methods of psycholexical research are limited. It reduces or eliminates the need to focus on a small number of personality descriptors to prevent participant fatigue. This reduces the effects of researcher bias in subsetting the terms to study. By circumventing the need for participant responses, NLP methods substantially reduce the costs and time associated with data collection and allow for sensitivity analyses of structure across a wide range of contexts. Though the data generated from language models is quite different from traditional participant-generated data, NLP data is generally sourced from multiple corpora of larger size, scope, and diversity than typical samples of survey-based adjective ratings. There are at least two relative disadvantages of the NLP approach. One is the inability to specify characteristics of the individuals who contributed to the source data. The second is a lack of evidence regarding the validity of language model-generated data for analyses of personality structure; the current work was designed to address this point.

**The current studies**

This work addresses two research questions. The first question is methodological in nature: is it feasible and valid to use publicly available, pre-trained NLP models in psycholexical studies of personality structure? Though modern language models excel at many diverse tasks (Wang & Komatsuzaki, 2021), it is not yet clear how well they can interpret the contextual cues and semantic relations specific to personality attributes. If NLP models are suited to the study of personality structure, they will yield results which are interpretable – though not necessarily familiar to personality psychologists. The second question, therefore, is how well does the structure evident in natural language compare to the substantial body of prior psycholexical research?

This latter question has implications for theory in personality structure research. As survey-based ratings have been used almost exclusively as the basis for previous structural research, it is unclear how well adjective ratings map the relations among personality descriptors. It may be that ratings data reflects semantic structure closely. This would be consistent with a strict interpretation of the lexical hypothesis (Saucier and Goldberg, 1996b; Wood, 2015) and would be supported by evidence that the correlational structure is highly similar between the two types of data (the traditional and NLP approaches). Though high levels of dissimilarity across the two sources of data are not likely, even moderate levels would point to the need for additional research to evaluate the causes of inconsistency. In essence, application of the NLP approach provides an opportunity to validate the ratings method in terms of capturing semantic structure.

Across three studies, we conduct multiple exploratory analyses of personality structure, mainly using a class of models known as transformer-based encoder-decoder models (described in Study 2). These analyses make use of several large sets of personality descriptors, including previously used sets with 435 terms (Goldberg, 1990; Saucier & Goldberg, 1996a) and 1,710 terms (Ashton, Lee, & Goldberg, 2004; Goldberg, 1982), and two sets that have not been considered previously (the 18,000 term set of Allport and Odbert [1936] and an 83 term set from Webster's Learner's Dictionary [Merriam-Webster, 2008]).

Given the exploratory nature of these analyses and the underlying techniques, neither hypotheses nor expected findings were pre-registered. It remains a priority however to situate our findings relative to prior evidence on the structure of personality descriptors specifically and personality models more broadly (including structural models based primarily on theory and/or empirical evidence using variables other than trait-descriptive adjectives). We expected an interpretable structure among the adjectives would emerge using NLP techniques and that this structure would largely reflect prior psycholexical research. To provide context for this, we summarize prior findings about personality structure based on traditional psycholexical approaches in Section 1 of the Supplemental Materials.

In Study 1, we reproduce the structure of the Big Five using existing data from Saucier and Goldberg (1996). In Study 2, we use the variables and analytic techniques of Study 1 with language model-generated data. Study 3 has three parts, each designed to evaluate different methodological decisions in the NLP approach.

## Study 1

The primary goal of this study is to reproduce the structure of the Big Five using one of the foundational sets of personality descriptors. This aim was motivated by the intention to directly compare results from traditional and NLP approaches in subsequent analyses; it was expected that this step would allow us to eliminate the possibility that differences between the two methods might derive from sources beyond the data-generating procedures (i.e., differences due to analytic methods or statistical software). We also sought to evaluate the evidence for alternative solutions (i.e., fewer factors) as a baseline for subsequent comparisons, even though they were expected to fit the data less well than the five-component solution.

The personality descriptor set used here is the 435-term set analyzed in Saucier and Goldberg (1996; dba "S&G 1996"). The rationale for this choice was based on several benefits, including: (a) the relatively large size and unstructured nature of the descriptor set, (b) the transparent presentation of prior results, making reanalysis feasible and straightforward, (c) the characteristics of the sample, including its large size, good mix of self- and peer-ratings, and its lasting influence on Big Five research, and (d) an expectation that these 435 terms are somewhat less biased than other subsets from the same era because they were empirically selected on the basis of familiarity in response to concerns about variable selection procedures (Block, 1995; Saucier & Goldberg, 1996a). More detailed discussion of this topic is provided in Section 2 of the Supplemental Materials; other adjective sets are evaluated in Study 3.

## Method

### Participants, Measures, and Procedure

The data were collected using 435 personality descriptors chosen from larger sets of terms on the basis of familiarity ratings by 25 raters (Saucier & Goldberg, 1996a). These raters were (mainly) undergraduates and law school students who were similar in age and education level to the 583 unique respondents who provided self-ratings (N = 507) and/or peer-ratings (N = 392). All of the terms were English and the data were collected using paper forms. Response options varied within the sample (7 or 8), ranging from "extremely inaccurate" to "extremely accurate". None of the ratings contained missing data.

*Analyses*

Upon receipt of the original data, the self- and peer-report ratings had already been ipsatized in order to address the differences in response options and to reduce the effect of differences in response style across raters. The first analytic step was to conduct a principal components analysis. We extracted 10 components to compare the first 10 eigenvalues against those reported in the manuscript. Then, using only the first 5 components, we rotated the loadings matrix to match the original analyses conducted in SPSS; this was varimax rotation, the default rotation method in the ROTATION subcommand of the FACTOR command. This rotation was done in R version 4.1.1 (R Core Team, 2021) using the GPArotation package (Bernaards & Jennrich, 2005) and following the method recommended by Weide and Beauducel (2019) to meet the SPSS-specific procedure.

To evaluate the evidence for alternative structures, two analytic approaches were considered. The first was the so-called "bass-ackwards" approach described by Goldberg (2006), as implemented in the psych package (Revelle, 2021) based on Waller (2007). The bass-ackwards approach extracts components sequentially and reports on the correlations between components across levels. In other words, the procedure begins by extracting one component, and then two components, reporting the correlation between the components in the one- and two-component solutions. This continues up to the maximum number of components to be extracted. We stopped at five, as we only sought to consider solutions that were more parsimonious than the Big Five reported by S&G 1996.

The second analytic approach to evaluate alternative structures was to consider the effect of rotations on interpretation of the dimensionality reduction. The effect of rotation was evaluated by comparing the congruence coefficients of components in unrotated and varimax-rotated solutions at levels of extraction between two and five. Tucker's congruence coefficients (Burt, 1948; Tucker, 1951) were used – here and in subsequent studies – following the recommendations of Lorenzo-Seva and Ten Berge (2006) such that values above .95 indicated identical components, values between .85 and .94 suggest fair similarity, and lower values suggest higher dissimilarity.

*Transparency and Openness*

The data, analysis code, and additional supplemental materials (for all studies) are available at osf.io/xm7hg. Approximate reproduction of the original data used here is also possible from the complete loadings matrix included in the original publication (Saucier & Goldberg, 1996). Given the aims, sample size determinations were not relevant. Principal components analyses were conducted using the psych package (Revelle, 2021) in R (R Core Team, 2021) in order to make use of the rotation methodologies in the GPArotation package (Bernaards & Jennrich, 2005). A discrepancy between the original results from SPSS and the results produced in R was resolved with code from Weide and Beauducel (2019).

## Results

Reproduction of the principal components analysis by S&G 1996 led to a perfect match to the loadings matrix reported in Table 2 of the original publication, with only three exceptions.[2] All of these appear to be typographical errors, as detailed in the Supplemental Materials. The eigenvalues for the first 10 components were also reproduced identically.

We also evaluated the evidence for alternative structural solutions in two ways. First, we evaluated the extent to which the 5 component varimax solution differed from solutions with fewer components. Following the bass-ackwards approach (Goldberg, 2006; Loehlin & Goldberg, 2014), we extracted components incrementally from 1 to 5 with varimax rotation in order to compare the correlations among components across levels of extraction (see Supplemental Figure 2). These results indicated that the extraction of additional components after 2 or 3 did not meaningfully alter the organization of content in the first 2 or 3; correlations across levels of extraction for these components were all greater than .90. Similar results were produced by identifying the congruence coefficients among the components in the varimax-rotated Big Five reported by S&G 1996 and the 2- and 3-component varimax solutions (see Supplemental Figures 3 and 4). For example, the 3-component solution was highly congruent with Agreeableness (.99), Extraversion (.97), and Conscientiousness (.94), respectively.

---

[2] Note that adjustments were needed to account for differences in the varimax rotation procedures used in the original SPSS output and current defaults in R. See the Supplemental Materials for details.

Second, we evaluated the effects of varimax rotation on the solutions. The effect of varimax rotation, by design, is to spread the proportion of variance explained by each component more evenly. In this case, the overall variance explained by the 5 component solution was 22.4%. The proportion of variance explained by each successive component in the unrotated solution it was 0.317, 0.259, 0.183, 0.131, and 0.100; in the varimax solution the proportions were 0.268, 0.250, 0.212, 0.143, and 0.120. About 15.4% of the variance explained by the first component was shifted to the third, fourth, and fifth components using varimax rotation. We also compared congruence coefficients between the unrotated and varimax rotated solutions for 2, 3, and 5-components (see Supplemental Figures 6 through 8). Rotation of the 2-component solution did not meaningfully alter the first (.97) or second (.95) components relative to the unrotated components, but the degree of congruence dropped such that all but one of the components in the 5-component solution (the second, Extraversion, at .97) were different between the unrotated and varimax-rotated solutions, and there were multiple high cross-loadings.

**Study 1 Discussion**

Given the perfect reproduction of the analyses reported in S&G 1996, we conclude that there are no differences between the statistical approach used here and prior exploratory work leading to identification of the Big Five. Reproduction of the previously-reported 5-factor solution in this study is only meaningful in terms of providing a foundation for subsequent studies.

The evidence for alternative structural solutions is relevant for subsequent studies in multiple ways. Recognition that the first 3 components, and especially the first 2, are highly robust to subsequent extraction provides further insight into the structure of the data. The question of whether 5 is better than 3, for example, can be simplified to the question of whether the fourth and fifth components are necessary or justified. Further, this finding creates an expectation that structural replicability is not "all or nothing," and that the first 2 (or 3) components are more likely to replicate than the last 2 (or 3). It is also interesting to note the influence of varimax rotation in making the third (Conscientiousness), fourth (Emotional Stability), and fifth (Intellect) components more prominent in the 5 component solution. It has the effect of increasing the variance explained by these components by 16%, 9%, and 15%, respectively, at the expense of the first component. This inflates the unrotated contribution of those components. When comparing across levels of extraction, the use of unrotated solutions is less complicated because the variable loadings and the amount of variance explained by each component do not change with the extraction of additional components as they do when rotating the solution at each level of extraction. For the sake of comparison in Study 2, we use the varimax solution(s) reproduced here from S&G 1996 and the unrotated solution.

**Study 2**

The aim of Study 2 was to evaluate the structure of the 435 terms in S&G 1996 based on NLP data. As described below, this was done using a language model with transformer architecture to generate vectorized representations of each term in the set of personality descriptors. We then compared the correlations among these model-generated vectors against the data from S&G 1996, including congruence analysis of the first 5 components.

As thousands of language models have recently been developed for generating vectorized representations of text, the nomenclature for describing them and their underlying procedures is not well established. Many of the available models have been designed using slightly different techniques and/or for different purposes, but they can be classified into similar groups. The models used in this and subsequent studies belong to a broader class of transformers sometimes referred to as contextualized learning models (CLMs), and these are distinct from several older classes of language models. For more information on the evolution of language models, see Zhang et al. (2020).

We focused on newer models because older models (e.g., word2vec, GloVe) are unable to account for contextual factors, such as those that enable disambiguation of polysemous words like "patient" (Yenicelik et al., 2020). More practically, contextual models were preferred because they allow for the development of queries that match the form and spirit of survey-based methods in lexical research. As previously discussed, the vectorized representation of each personality descriptor is functionally equivalent to the vectors produced based on self- or other-report ratings of terms. For a recent primer on working with transformer architecture models, see McCaffrey (2021) and Cistac et al. (2019).

The process of developing new models de novo requires considerable time and expense, but thousands of existing, "pre-trained" models are available open source through a library (and API) hosted by Hugging Face (huggingface.co). All the models considered in this work (Studies 2 and 3) were accessible through Hugging Face at the time of submission; however this resource is frequently updated with new models and versions. No attempt was made to exhaustively compare all possible options. We begin by using only one model in Study 2, but many additional models are evaluated in Study 3.

## Method

Three sets of materials are needed to conduct the analyses in Study 2. Specifically, these include a (1) *model* that uses (2) *queries* to generate contextualized vector representations for each of our (3) *terms*. Given our aims to compare structure against the results of Study 1, the terms are the 435 from S&G 1996. The processes of selecting a model and writing queries both required (and allowed for) considerable flexibility in study design. Given the exploratory nature of this work, numerous models and queries were considered before settling on the options reported in the current study. As this exploration necessarily increased the likelihood of bias in our results due to "researcher degrees of freedom" in study design choices (Simmons et al., 2011), the effects of different choices are presented as sensitivity analyses in Study 3.

*Model selection.* Study 2 used a model known as DeBERTa (Decoding-enhanced BERT with disentangled attention; He et al., 2021), and more specifically, a version of the model that was fine-tuned to maximize performance on the multi-genre natural language inference task (Williams et al., 2017). DeBERTa was chosen because it was the top-performing model on several benchmark tasks, though it has since been superseded. It is common for language models to be developed/trained to perform well on one or more specific tasks (Han et al., 2021), and this information is tracked publicly using benchmarks like GLUE (the General Language

Understanding Evaluation) and the more difficult SuperGLUE (super.gluebenchmark.com;

Wang & Komatsuzaki, 2021), an index of 8 language understanding tasks. At the time of

submission, DeBERTa is one of three models outperforming the human baseline estimate, on

average, across all 8 tasks. DeBERTa was pre-trained using deduplicated text (Lee et al., 2021)

from Wikipedia (Wikimedia downloads), BookCorpus (11,038 books from 16 genres; Zhu et al.,

2015), OpenWebText (38GB of Reddit submissions, Gokaslan & Cohen, 2019), and a subset of

CommonCrawl data (commoncrawl.org; Trinh & Le, 2018). For additional details about development of the model, see He et al. (2021) and github.com/microsoft/DeBERTa.

*Query writing and selection.* We used a novel method for generating vectorized word

representations developed specifically for the current project, though many strategies are

possible using queries (sentences, in this case) as inputs to the pretrained models. A simple

example would be:

```
          "She has a [TERM] personality."
```

In general, contextualized learning models work by initially interpreting each query as a

sequence of tokens. Many methods exist for tokenizing queries, ranging from those that split text

into strings of individual characters to those that use whole words or phrases. In the example

above, if the term is "condescending", the query may be tokenized into sub-words as follows:

```
 [CLS] [She] [ has] [ a] [ condesc] [ending] [ personality] [.] [SEP]
```

Note that spaces are carried along at the beginning of each word, if applicable, and the word

"condescending" is split in two. In DeBERTa (and similar models) every sequence begins with the special CLS (classification) token. The model uses this to encode general information about the entire sequence, and this is the token that attends to other tokens the most often. Among many other characteristics, for example, the CLS token might encode the extent to which the sequence is written in King James English or internet slang. For some downstream use cases, like sentiment analysis, only the CLS token is used, and the other token representations are discarded. The [SEP] token is applied to the end of a sentence. It is important to signal separation of sequences and, for some models, to predict whether sentences occur in the correct order. Tokens can also be inserted or replaced with blanks to be filled in by the model during processing based on the context of the surrounding text. These blanks are known as mask tokens (discussed further below). See Rust et al. (2020) and Cistac et al. (2019) for more information on tokenization, noting that models vary with respect to methods and strategies for tokenizing.

In addition to each query being interpreted as a sequence of tokens, each token is represented as an initial static vector. With most models (including those used here), these initial vector representations are informed by all uses of the token in the underlying corpus. The models account for context by transforming this initial vector with weights that capture syntactic and semantic features. One example of these features is the position of the token (word) in the query (sentence), but most transformations rely upon more complex features. The resulting weighted vectors are the output of the pretrained models for each token of each query.

Though there are many possible means of using the series of token vectors produced for each query, we used a strategy that approximates the approach of collecting adjective ratings. Specifically, we used the representations (i.e., the vectors) of special fill-in-the-blank mask tokens. This strategy loads personality-relevant information from the word of interest onto one or more masked tokens, which the model recognizes as a "wild-card." Results of the current study are based on the following query:

"Those close to me say I have a [MASK][MASK] and [TERM] personality."

Recall that the terms used here are the 435 from Saucier and Goldberg (1996); each term would be represented by a separate query, and each query would be input to the model as a sequence of tokens. With this query format, the model output used for each term was the average of the two [MASK] token vectors. This is advantageous because many words require two mask tokens to represent. Using only a single mask token biases the model towards short, common words. The model is attempting to fill in the [MASK] token and thus must figure out what words likely occur there, given the context. Semantic and positional information about the missing token is stored as a 1,024-dimensional vector that subsequent layers of the model will unpack to be a probability distribution over every token in the vocabulary (about 30,000 for this model). Dimensionality of the encoding vector (i.e., length of 1,024 in DeBERTa) is a tradeoff between computational resources and representational complexity; some models use different lengths (e.g., 512, 768, 2,048). In other words, the context of each query – one version for all 435 terms – prompts the model to determine what tokens would fill the [MASK] slot and represent it with a vector that encodes its content and position relative to other tokens based on a lower level of dimensionality than is contained in the full space of the vocabulary tokens.

Note that it is also possible to combine the results from multiple queries to reduce the idiosyncratic noise of query syntax, and this may be recommended in cases where different queries offer complementary strengths and weaknesses. More discussion of this topic is provided in Study 3, where we also evaluate the effect of using alternative queries.

*Analyses*

The model-generated data are accessible at [osf.io/xm7hg](osf.io/xm7hg). These data were generated by looping the query through the model for each of the 435 terms. Analyses of the output followed the same steps as those used in Study 1. Of note, the form of the resulting data was approximately equivalent to the data frame underlying the results reported in S&G 1996; both are two dimensional matrices with 435 variable columns. The S&G 1996 data frame contains vectors with 899 dimensions (i.e., rows) based on self- and other-ratings. The NLP data frame contains vectors with the coordinates of each term in the n-dimensional space of the DeBERTa language model (1,024 rows). To compare structure across the data sources, we generated heat maps of the correlation matrices, and conducted congruence analyses to show the degree of similarity between principal components when extracting 3 or 5 components, with and without (varimax) rotation. The analyses in Study 2 are documented in Section 4 of the Supplemental Materials.

## Results

Correlations among the 435 personality descriptor variables are shown in Figure 1. For comparison, the correlations among the variables in S&G 1996 are shown on the left side of the image and correlations from the DeBERTa model are shown on the right. Note that the variables are identically ordered in both matrices using the hierarchical clustering algorithm in R (hclust; R Core Team, 2021) for easier visual comparison. As the figure indicates, the magnitude of associations was much greater in the DeBERTa data, but the direction of the associations was highly similar across data sources. Comparison of the mean and median absolute values of the lower triangles of the two matrices confirmed the differences in magnitudes ($m$=0.086 and $Mdn$=0.069 for S&G 1996; $m$=0.248 and $Mdn$=0.232 for DeBERTa).

**Figure 1**

*Correlations among all 435 variables ordered by hierarchical clustering of the DeBERTa data*

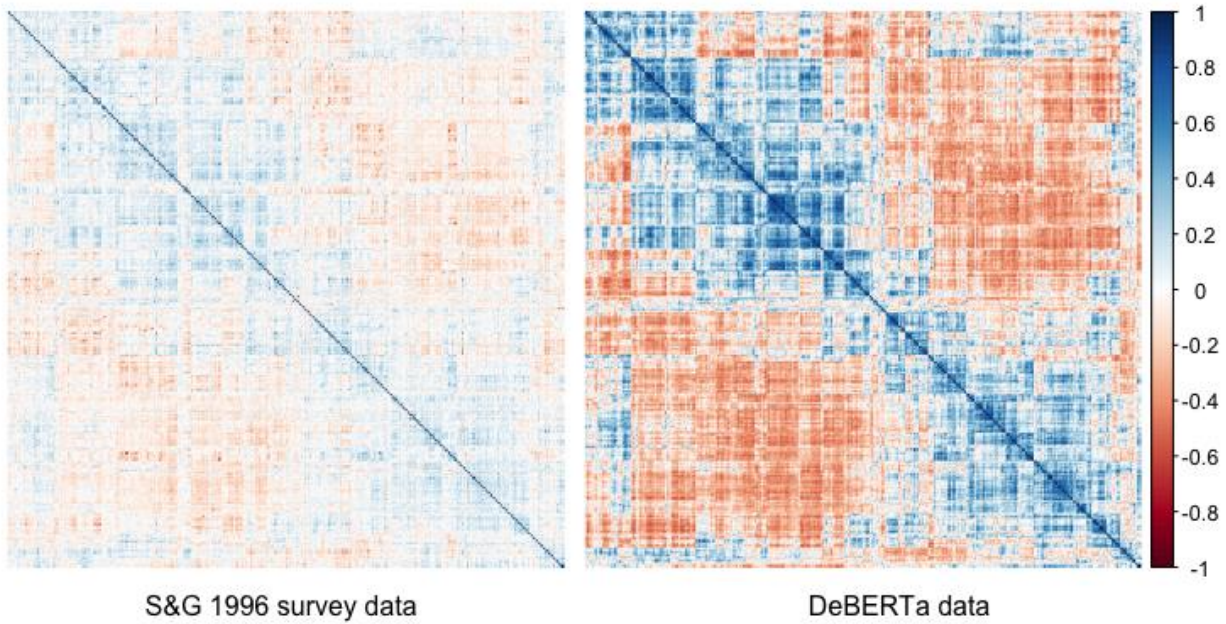

The same matrices shown in Figure 1 have been reorganized in Figure 2, this time

Analyses of the directional consistency from the S&G 1996 correlations to the DeBERTa

correlations showed that 75.1% of the 94,395 pairwise correlations were directionally consistent

across matrices (i.e., the correlation was less than 0 in both matrices or greater than 0 in both). Of

those that were directionally inconsistent, 82% had confidence intervals overlapping with zero in

the S&G 1996 correlation matrix (at $p<.01$). Only 4.5% of all 94,395 correlations were

statistically significant in S&G 1996 ($p<.01$) and had a different sign in the DeBERTa

correlation matrix. Congruence among the correlations for the statistically significant values in

S&G 1996 was .79.

The same matrices shown in Figure 1 have been reorganized in Figure 2, this time matching the ordering of variables by component loading in S&G 1996 (the original Big Five organization). Note that the directional similarity appears more consistent for the first three components, in the upper left corners of the matrices, than for the last two components in the lower right corners. In Figure 3, the same matrices are reorganized again to match the component loadings of the terms based on the unrotated 5 component solution for the DeBERTa data. This image shows the relative strength of the components – the first three are clearly visible, the fourth is discernible but small, and the fifth is not visible. The first component was about as large as the second and third components combined. The same pattern of associations was evident in the survey-based ratings data despite the lower magnitude correlations.

Figure 4 shows the congruence coefficients between the varimax rotated 5-component solution from the DeBERTa and the Big Five (in the upper matrix) and the Big Three (three varimax rotated components from S&G 1996; the lower matrix). Agreeableness, Extraversion, and Conscientiousness from the Big Five were recovered well (congruences between .79 and .85 on the diagonal). Neuroticism and Intellect were not well recovered and showed higher congruences with other factors. In the lower matrix, the first 3 components were recovered as clearly as in the Big Five case, and the remaining two DeBERTa components were also moderately congruent with the first 3 components. In both cases (the Big Five and the Big Three), the strongest congruences with the varimax-rotated DeBERTa data was with the third component.

**Figure 2**

*Correlations among all 435 variables ordered by component loading in S&G 1996*

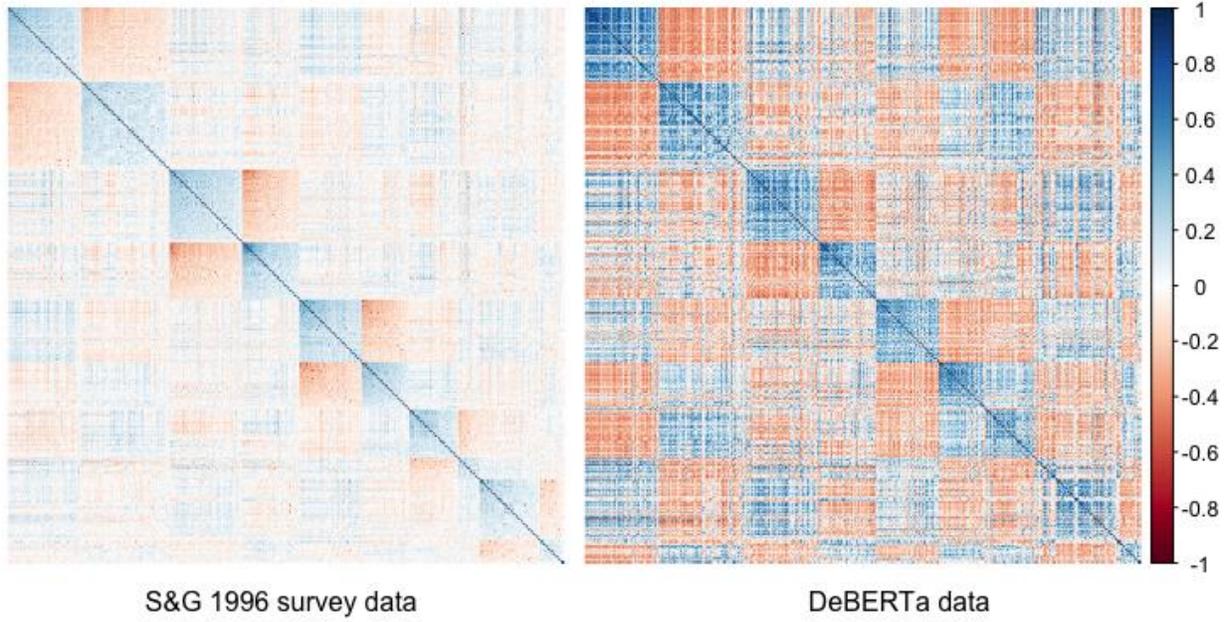

**Figure 3**

*Correlations among all 435 variables ordered by component loading in the DeBERTa data*

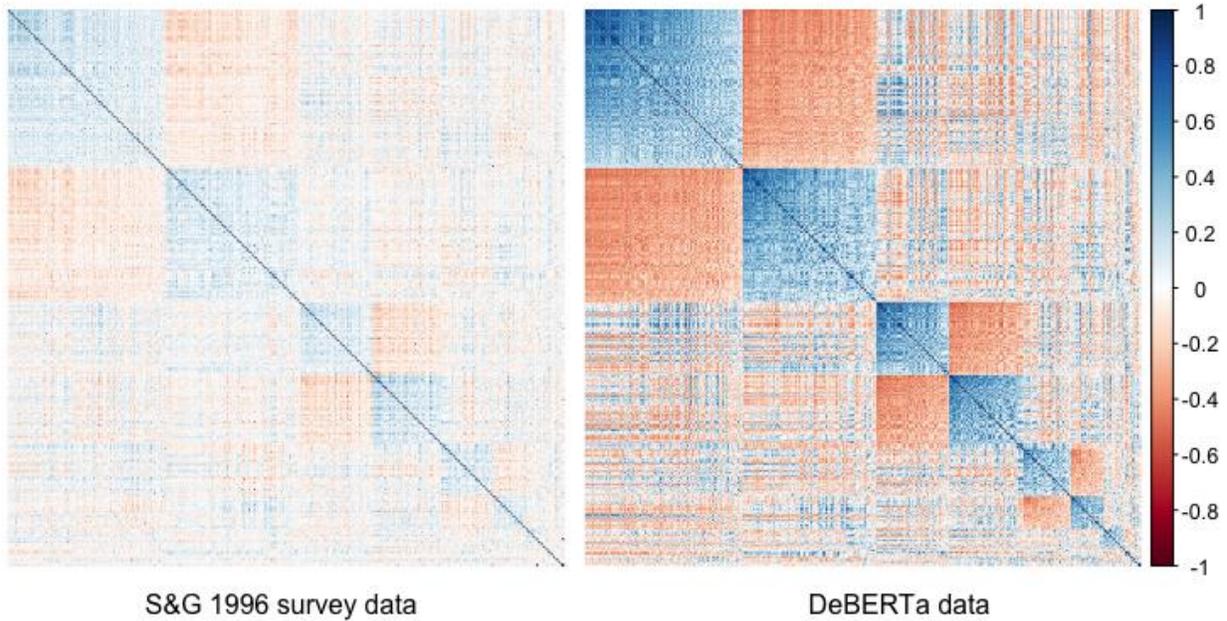

**Figure 4**

*Congruence coefficients between 5 and 3 varimax-rotated components from S&G 1996 data and*

*5 varimax-rotated components from DeBERTa data*

**Figure 5**

*Congruence coefficients between 5 unrotated components from S&G 1996 data and 5 unrotated*

*components from DeBERTa data*

Figure 5 shows the congruence coefficients more clearly, using the 5 component unrotated solutions for the S&G 1996 data and the DeBERTa data. The congruences were high for the first 3 components, but the last 2 components were not well recovered. Table 1 shows the 8 highest loading terms for each component in the DeBERTa data. The terms in the first component were highly consistent with Agreeableness/Affiliation. The terms for the second and third components were also well-aligned, though not exactly, with operationalizations for Extraversion/Dynamism and Conscientiousness/Order, respectively. The fourth and fifth components were even less well aligned with Neuroticism and Intellect/Openness. Terms on the fourth component reflected concepts in both Neuroticism (introspective, emotional) and Intellect (philosophical, complex). The fifth component was difficult to interpret and small (only 7 of the 435 terms had primary loadings on this component).

The last step of Study 2 involved analyses of the differences between the data sources at the level of individual terms. This was done by identifying the nearest and furthest neighbors of each term in each of the correlation matrices. For example, the nearest neighbors of "fearful" in the S&G 1996 results were "cowardly", "fretful", and "gullible", while the nearest neighbors using DeBERTa data were "insecure", "cowardly", and "fretful". The correlations across data sources of each term's correlation with all other terms were generally high ($m$=0.70; $sd$=0.15; $Mdn$=0.74); 87% (of 435) are correlated at .60 or higher. When the terms were sorted by Big Five trait using the S&G 1996 results, the magnitudes of mean correlations for the 10 most

**Table 1**

*Component loadings for the 8 terms with the highest primary loadings (absolute value) based on*

*the unrotated 5-component solution using DeBERTa data.*

| Term | PC1 | PC2 | PC3 | PC4 | PC5 |
|------|-----|-----|-----|-----|-----|
| amiable | **0.83** | 0.04 | -0.29 | -0.23 | 0.20 |
| sincere | **0.81** | -0.07 | -0.09 | -0.10 | 0.20 |
| generous | **0.80** | 0.11 | -0.21 | -0.09 | **0.35** |
| genial | **0.78** | 0.06 | -0.29 | -0.25 | 0.15 |
| kind | **0.78** | -0.04 | -0.13 | -0.07 | **0.30** |
| trustful | **0.78** | -0.19 | -0.08 | -0.17 | **0.34** |
| intelligent | **0.77** | 0.24 | 0.24 | 0.01 | 0.18 |
| courteous | **0.76** | -0.23 | -0.06 | -0.25 | **0.36** |
| impetuous | -0.03 | **0.84** | -0.01 | -0.02 | -0.19 |
| excitable | 0.13 | **0.80** | -0.22 | -0.06 | -0.12 |
| rambunctious | 0.07 | **0.80** | -0.22 | -0.07 | -0.24 |
| sedate | 0.26 | **-0.80** | -0.12 | 0.01 | -0.14 |
| restrained | 0.27 | **-0.78** | 0.10 | 0.02 | -0.12 |
| reserved | 0.07 | **-0.77** | 0.07 | 0.18 | -0.12 |
| unexcitable | -0.16 | **-0.77** | 0.04 | 0.01 | **-0.33** |
| zealous | 0.12 | **0.76** | 0.20 | -0.10 | -0.21 |
| lax | 0.06 | -0.11 | **-0.76** | -0.13 | -0.27 |
| exacting | -0.29 | -0.14 | **0.74** | -0.02 | -0.05 |
| gullible | -0.14 | 0.06 | **-0.69** | -0.05 | 0.18 |
| strict | -0.25 | **-0.39** | **0.69** | -0.14 | -0.12 |
| decisive | 0.13 | 0.17 | **0.68** | -0.15 | -0.25 |
| suggestible | -0.09 | 0.16 | **-0.66** | 0.06 | 0.14 |
| stern | -0.23 | **-0.34** | **0.65** | -0.04 | -0.17 |
| naive | 0.22 | 0.06 | **-0.64** | -0.03 | -0.06 |
| introspective | 0.08 | **-0.49** | 0.11 | **0.61** | 0.06 |
| philosophical | 0.21 | -0.10 | 0.18 | **0.58** | -0.03 |
| emotional | 0.25 | **0.40** | 0.01 | **0.55** | 0.18 |
| complex | 0.22 | 0.27 | 0.22 | **0.53** | 0.00 |
| boastful | **-0.35** | **0.50** | -0.01 | **-0.52** | -0.04 |
| contemplative | 0.01 | **-0.49** | -0.03 | **0.52** | 0.02 |
| melancholic | **-0.32** | **-0.41** | -0.08 | **0.52** | 0.19 |
| subjective | -0.10 | 0.18 | -0.20 | **0.51** | -0.17 |
| nonchalant | 0.23 | -0.24 | **-0.38** | -0.25 | **-0.55** |
| concise | **0.32** | -0.25 | **0.36** | -0.12 | **-0.47** |
| vague | -0.25 | -0.21 | **-0.43** | 0.29 | **-0.46** |
| direct | **0.30** | 0.02 | **0.33** | **-0.40** | **-0.43** |
| autonomous | 0.11 | 0.24 | 0.17 | 0.27 | **-0.38** |
| weariless | **-0.31** | **-0.36** | -0.21 | 0.02 | **-0.37** |
| nonreligious | 0.04 | -0.16 | -0.18 | -0.01 | **-0.21** |

*Note:* Only 7 variables have primary loadings on the fifth component.

highly loaded terms mirrored the ordering of component extraction (Agreeableness: 0.82; Extraversion: 0.78; Conscientiousness: 0.75; Neuroticism: 0.69; Intellect: 0.65). As the nearest and furthest neighbors are essentially synonyms and antonyms, respectively, there were some cases that suggested confusion about meaning among the survey respondents ("animated", "gossipy", and "meek" identified as nearest neighbors for "transparent") or the language model ("passionless", "dull", and "lethargic" identified as nearest neighbors for "weariless"). However, the differences were subtle in most cases, as expected given the generally high, positive correlations. Appendix B of the Supplement provides a listing of these results for both data sources.

Results of the nearest and furthest neighbors analysis also demonstrated that the language model is inferring the appropriate, personality-specific meaning of terms that have multiple definitions. For example, near neighbors of "transparent" were candid, frank, and truthful, and "cold" was most correlated with aloof, detached, and unfriendly.

## Study 2 Discussion

Direct comparison of the survey-based ratings and the language model output led to several insights. First, the high directional consistency in correlational structures provides substantial evidence that the signal provided in these self- and other-ratings was largely a reflection of semantic structure among the terms. Relatedly, the correlations among ratings of the terms captured a large amount of idiosyncratic noise, diluting the magnitudes of the correlations; the evidence for this is the higher magnitude correlations among terms in the DeBERTa data as these necessarily reflect the semantic associations more directly. Additional insight follows from the difficulty of recovering all of the Big Five components. Only the first 3 components were congruent, and the structure of these (unrotated) components in the DeBERTa data appeared to reflect Affiliation, Dynamism, and Order (i.e., distinct variations on Agreeableness, Extraversion, and Conscientiousness). Organization of the correlation matrices based on the structure of DeBERTa data, which has more signal than the survey data, make it clear that the fourth and fifth components — Neuroticism and Intellect — contain little content and are hard to interpret. Further consideration of the theoretical consequences is provided in the General Discussion, following Study 3.

### Study 3

Study 3 was designed to evaluate the effects of the main methodological decisions in Study 2 relating to the selection of (1) personality descriptors, (2) query, and (3) model. Each of these was evaluated with a separate sub-study. In Study 3.1, we use three alternative sets of descriptors with the same query and model (DeBERTa). In Study 3.2, we use the 435 terms from S&G 1996 with DeBERTa and a range of different queries. In Study 3.3, we use many different models without varying the query or set of descriptors. Given the use of this novel methodology, our aims were exploratory. We sought to demonstrate, in general, the relative influence of each methodological aspect, and to identify specific topics for future research.

In all 3 sub-studies, congruence analyses are used to evaluate the similarity between the first 5 unrotated components in the ratings data from S&G 1996 and the language model(s) data. Given the lack of clarity provided by the use of varimax rotation in Studies 1 and 2, we did not compare varimax rotated solutions in Study 3 except as an additional analysis in sub-study 3. We also created heat maps of the correlation matrices generated in each of the sub-studies to aid with visualization of the structure. See osf.io/xm7hg for the data, code, and supplemental materials.

## 3.1 Methods

*Materials*

The model and query used in Study 3.1 were identical to those used in Study 2. The model was a large version of DeBERTa fine-tuned on the multi-genre natural language inference task. The query was "Those close to me say I have a [MASK][MASK] and [TERM] personality." Thus, the only distinction from Study 2 was to consider the effect of using personality descriptor sets that were different from the 435 terms in S&G 1996. We chose 3 overlapping sets of various sizes. The first of these was the complete set of 17,913 terms described by Allport and Odbert (A&O18k; 1936). Note that this reference claimed 17,953 terms in total, but many duplicates were identified in their list (see Section 5 of the Supplement for specifics and further details about the computational requirements of working with this adjective set). The second set was the 1,710 terms described by Goldberg (G&N1710; 1982), developed in collaboration with Norman, who had previously identified a longer list (2,797 terms; Norman, 1967). The third was a smaller set of only 83 descriptors from a list posted online by Webster's dictionary (Webster83; Merriam-Webster, 2008). These terms are taken from Merriam-Webster's Learner's Dictionary, one of many learners' dictionaries designed for language students. The Merriam-Webster version contains 3,000 core vocabulary words that Merriam-Webster editors deemed to be among the most essential English terms. The 83 words used here are taken from the "Personality Types" category.

## 3.1 Results

Results of the congruence analyses are shown in Figure 6. These congruences show the relationships among the loadings based on the unrotated components using DeBERTa data for the 435 terms from S&G 1996 and for each of the 3 additional descriptor sets. Note that the congruences are based only on the overlapping terms, though the loadings are based on the structure of each set of terms independently. The smallest set was most congruent with the terms from S&G 1996; these are the 70 overlapping terms from Webster83. Components 1 through 4 were nearly identical and component 5 was moderately well recovered (.59). The results were similar for the 434 overlapping terms from the G&N1710, but the congruence was higher for the fifth component (.67). The A&O18k set contains all 435 terms in S&G 1996 but the components suggested a different structure. The first 3 components were recovered again, but the fourth A&O18k component was split across the third and fifth components from S&G 1996. The fifth component was not congruent with any of the components in the Study 2 results.

**Figure 6**

*Congruence coefficients between the first five unrotated components from DeBERTa output using the S&G 1996 terms (Study 2 results) and each of the 3 new sets considered in Study 3.*

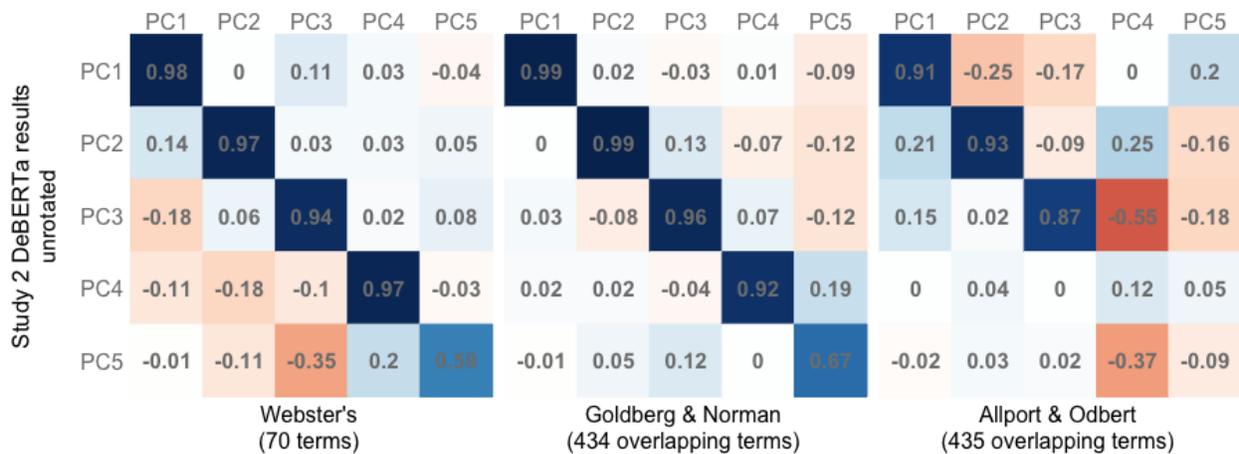

Figure 7 shows the heat map of the correlations among terms in A&O18k after organization with the same hierarchical clustering approach used in Study 2. Some structure was evident, as indicated by the darker blue squares down the diagonal, but there was also a larger proportion of low magnitude correlations than in the DeBERTa data from Study 2. The associations were generally stronger for the G&N1710 descriptors, and stronger still for the Webster83 set (see Supplemental Figures 17 and 18).

## 3.2 Methods

*Materials*

The model and set of personality descriptors used in Study 3.2 were identical to those used in Study 2. The model was a large version of DeBERTa fine-tuned on the multi-genre natural language inference task. The descriptor set was the 435 terms used in S&G 1996. In this sub-study, we sought to compare queries that use the personality descriptors in different ways. As changes to the semantic structure of a query has the potential to alter the associations among terms in the model-generated data (and the resulting component structure), we tested 8 different queries (listed below), each drawing upon distinct features of personality evaluation in the natural language.

Query 1: "`I tend to be [MASK][MASK] and [TERM].`"
Query 2: "`Those close to me say I have a [MASK][MASK] and [TERM] personality.`"
Query 3: "`The girl's disposition became more [MASK][MASK] and [TERM] as the years passed.`"
Query 4: "`My inlaws seem like [MASK][MASK] and [TERM] folks.`"
Query 5: "`Met this guy and he's so [MASK] and [TERM]. You would not believe!`"
Query 6: "`My arch enemy's personality: [MASK], [MASK] and [TERM].`"

**Figure 7**

*Correlations among the 17,913 terms in the Allport & Odbert set in DeBERTa data after*

*hierarchical clustering.*

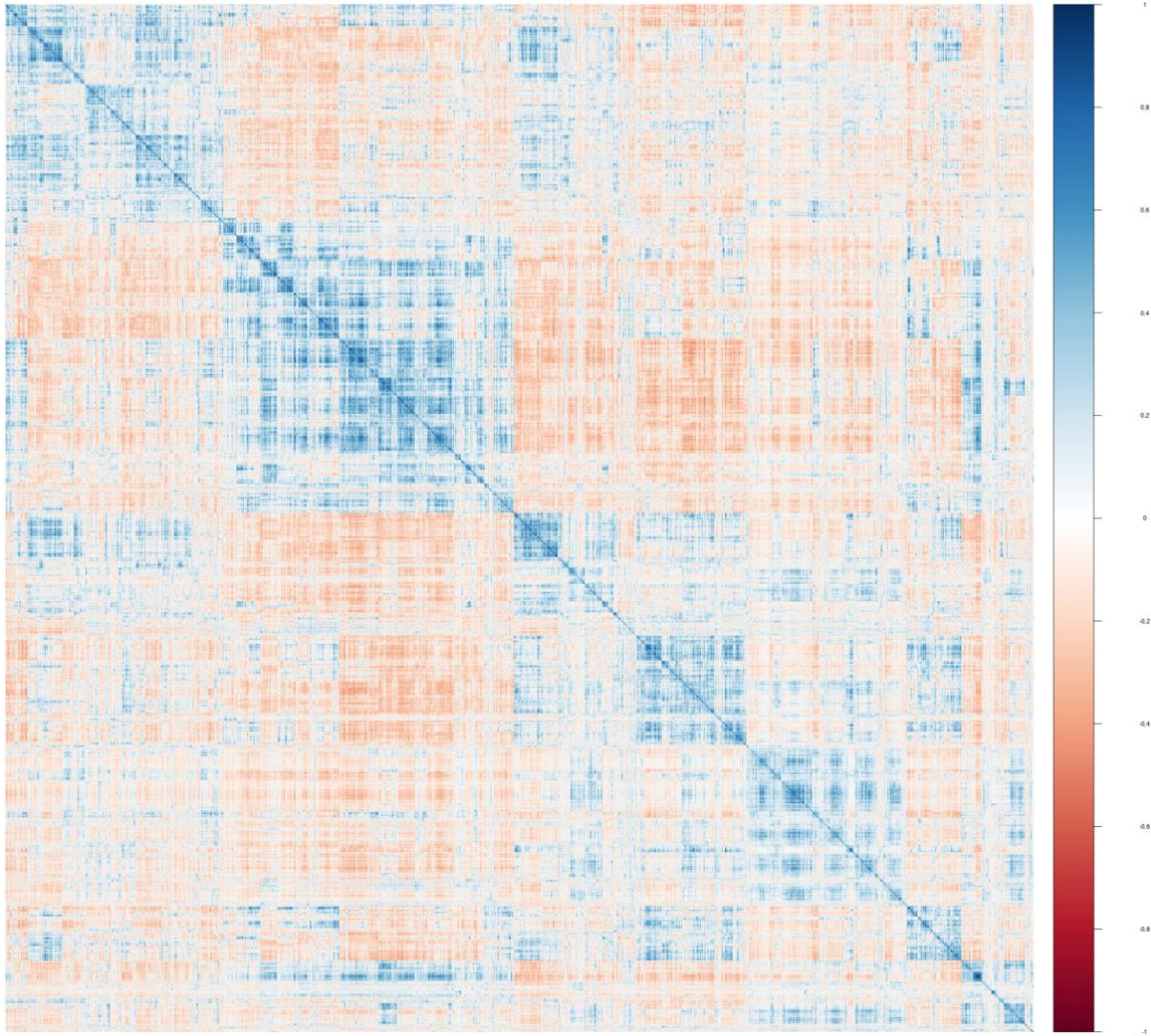

Query 7: "`A woman of contrasts: at times [MASK][MASK][MASK][MASK]`
`and at others perfectly [TERM]."`

Query 8: "`When he felt most authentic he would adopt an`
`[MASK][MASK] and [TERM] persona."`

Note that Query 2 is the same query used in Study 2 and the other sub-studies of Study 3. The queries are designed to be varied across type of relationship, intensity and grammar.

Commentary about each query is provided in Section 5.2 of the Supplement.

**Figure 8**

*Correlations among all 435 variables ordered by hierarchical clustering of the DeBERTa data*

*using all queries combined.*

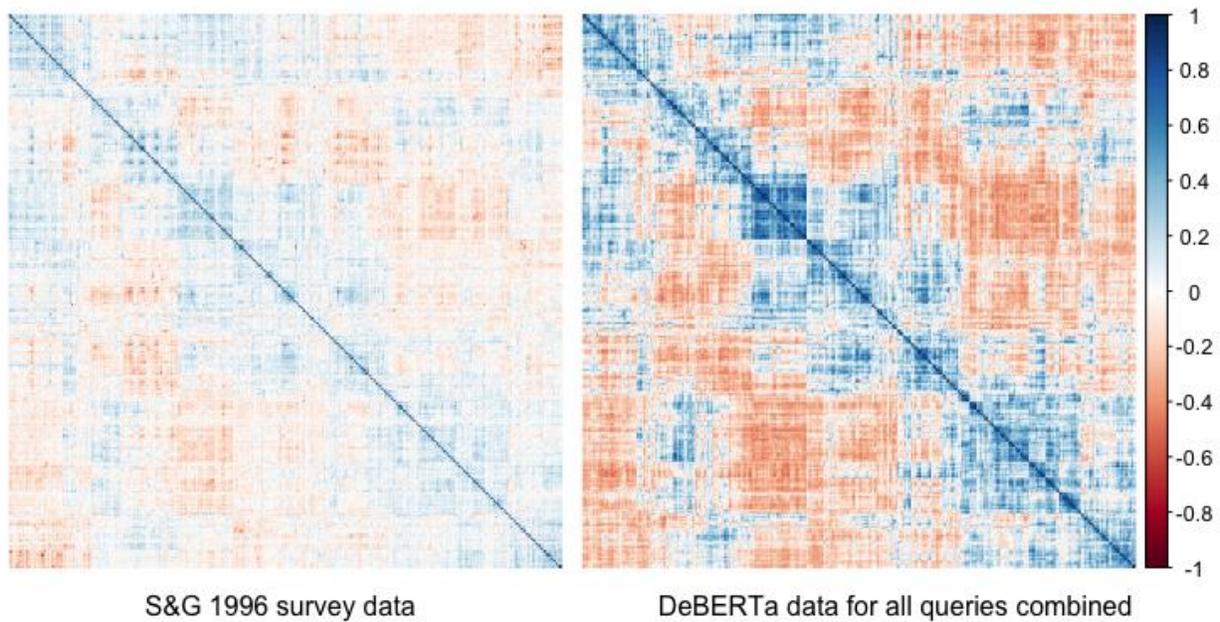

**3.2 Results**

Figure 8 compares the correlations for the combined set of queries and the survey-based ratings from S&G 1996; in this case, both matrices were organized according to the hierarchical clustering of the 8 queries combined. As in Study 2, the structure was highly similar across the two data collection methods, and the language modeling output provided a stronger signal. The mean and median absolute values of the lower triangles of the combined queries was .209 and .19 respectively; as reported previously, the mean and median for the S&G 1996 ratings data was .086 and .069, respectively. Of the correlations that were statistically significant in S&G 1996, only 3.0% had a different sign in the combined queries correlation matrix – a slight improvement relative to the use of only 1 query in Study 2.

Figure 9 extends this comparison with the S&G 1996 survey data by showing congruence coefficients among the first 10 unrotated components from each source. Congruence among the first 3 components was high (.93, .83, and .81). With the exception of modest congruence on the fifth component (.51), the remaining congruence values are less than .4, however several values between .2 and .4 suggest there was some shared signal among the later components.

Results for the congruence analyses among the 8 queries were based on the extent of similarity between each of the queries relative to all of them combined (see Supplemental Figure 21). In 6 of the 8 queries, the first 2 components were congruent at .89 or higher. For 7 of the 8 queries, the third component was congruent at .83 or higher. Only Query 8 deviated from the combined set on the first 3 components, and this was due to high congruence among the second and third components resulting from that query. The content in the fourth and fifth components was much less consistent than the first 3. Several of the queries (Queries 1, 2, 3, 4, and 6) showed high overlap between these components and/or reverse orientation of the content. This

**Figure 9**

*Congruence coefficients between each of the 8 queries independently and the combination of all 8 together.*

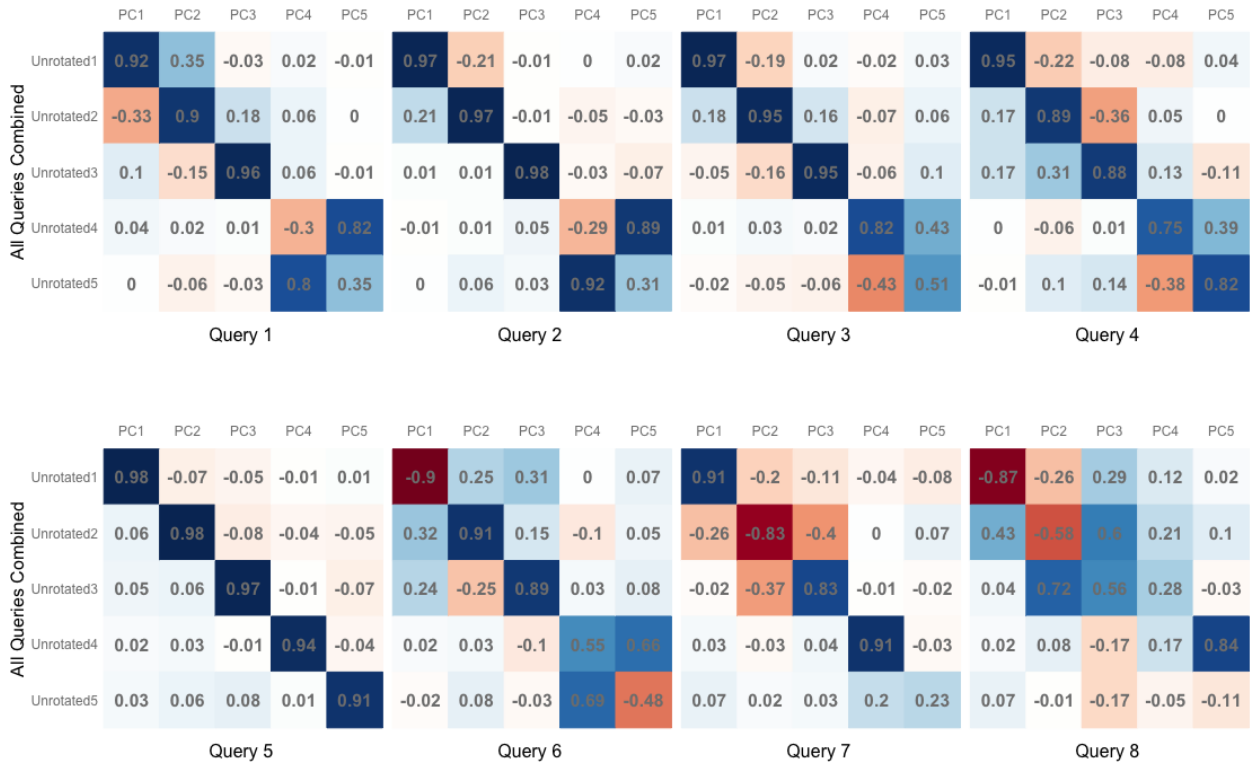

was further supported by follow up analyses of the congruences between the fourth and fifth components of all pairs of queries, and of the percentage of variance explained by the components of each query. The mean congruence between the fourth and fifth components across pairs of queries was .45. For most queries, both the fourth and fifth components explain a small proportion of the variance: mean variance for the first component was 20.0% across all 8 queries, 12.6% for the second, 9.3% for the third, 5.4% for the fourth, and 4.5% for the fifth. See Supplemental Tables 4 and 5 for more information.

Examination of the correlations suggests that the magnitude of associations among terms did vary as a function of query. To a lesser extent, there was also some evidence of directional changes in the correlations among terms across queries. Among all queries, the most clear and parsimonious structure emerged from the queries about in-laws (Query 4) and the "girl's disposition... as the years passed" (Query 3); by contrast, Queries 8, 5, and 1 produced more complicated structure. (The relevant heat maps for comparing correlations across queries can be found in Supplemental Figure 22.)

## 3.3 Methods

*Materials*

The personality descriptors and query used in Study 3.3 were identical to those used in Study 2. In this case, we used 17 alternatives to the DeBERTa model implemented in the prior analyses. These additional models were chosen to compare different versions of DeBERTa to one another and to compare DeBERTa against other models, including several that were optimized for specific tasks or use cases. Detailed descriptions and links to more information for each model are provided in Section 5.3 of the Supplementary Materials.

Among the different versions of DeBERTa, we compare the version used in all previous analyses – the large model fine-tuned with the multi-genre natural language interference task (*DeBERTa-l-mnli*) – against versions that are larger (i.e., trained with more parameters and layers, and larger vocabularies; *DeBERTa-xl-mnli*), and/or have not been tuned with MNLI (*DeBERTa-l, DeBERTa-xl*). We also compared the smaller base version of DeBERTa (*DeBERTa-b*).

As mentioned in Study 2, DeBERTa was initially chosen because it is among the best performing models of a type that is particularly useful for the current application. The model architecture for DeBERTa was developed by Microsoft (He et al., 2021) and is based on predecessor models developed by Google (BERT; Devlin et al., 2019), Facebook (RoBERTa; Liu et al., 2019), and others in the broader class of models known as transformers. Given this history, our comparisons included a large version of RoBERTa (*RoBERTa-l*), and multiple versions of BERT. These were *BERT-l* (large), *BERT whole word* (trained using masked tokens that span entire words), *BERT mobile* (designed for use in environments with limited computational resources like mobile devices), *BERT-of-Theseus* (a compressed model, like mobile, that was developed by iteratively compressing submodules of the model), *spanBERT* (a version pretrained for better handling of long spans of text), and *BERT clinical* (trained using text from electronic health records).

The remaining comparisons were made using models that employed specialized strategies for one or more aspect of the language modeling: *BART* is a transformer variation that was trained to reconstruct text that has been intentionally corrupted; *MPNet* permutes mask token handling in a text sequence; *Muppet* is fine-tuned for performance on 50 language tasks in the domains of classification, common-sense reasoning, machine reading comprehension, and summarization; the *Toxic* model was developed to classify online content in terms of toxicity; and the *Tweet* model was trained on a corpora with 850 million tweets. Finally, we included a cross-lingual language model that has been trained to classify across 100 languages – *XLM*.

## 3.3 Results

Comparison of the correlations from all 18 models combined and the S&G 1996 survey ratings, presented side-by-side as heat maps in Supplemental Figure 24, are consistent with the results of the prior studies – the direction of the correlations are highly similar across data sources and the magnitudes are much higher for the language models.

The results of congruence analyses comparing each model to the combined data from all models are shown in Supplemental Figures 25 and 26. These show high consistency among the models that perform well on other NLP tasks, including DeBERTa-l-mnli, DeBERTa-xl-mnli, DeBERTa-l, DeBERTa-xl, RoBERTa-l, and BERT-l. Variations on these models that are smaller (DeBERTa-base) or use different technical strategies for pre-training or fine-tuning (BERT whole word, BART, spanBERT, MPNet, muppet, XLM) deviate from the top performing models in that they match the combined model output on only 3 or 4 components, often with slightly lower congruence. Of the remaining models, only Tweet performs moderately well. BERT-clinical, BERT-mobile, BERT-of-Theseus, and Toxic match the other models well only on the first component.

Consistent with prior analyses, heat maps were generated for each of the models independently (Supplemental Figures 30 and 31). These produced similar findings to the congruence analyses comparing the models to one another – the largest and most modern models produced stronger associations among terms and clearer structure. These figures also show that the least congruent models are producing correlations that are noticeably weaker (BERT-Clinical) or unidimensional (Toxic).

Figures 10 and 11 show congruences between each language model and the Big Five from the S&G 1996 ratings (note that these are the varimax rotated Big Five components). These results show that model selection influences the resulting structure considerably. The large and extra-large versions of DeBERTa and RoBERTa recover the first 3 components of the S&G results fairly well and at similar levels of magnitude across models (Agreeableness [.73-.79], Extraversion [.73-.79], and Conscientiousness [.61-.68]). BART and Tweet do nearly as well

**Figure 10**

*Congruence coefficients between the first 9 (of 18) models and the S&G 1996 survey components*

*(Big Five).*

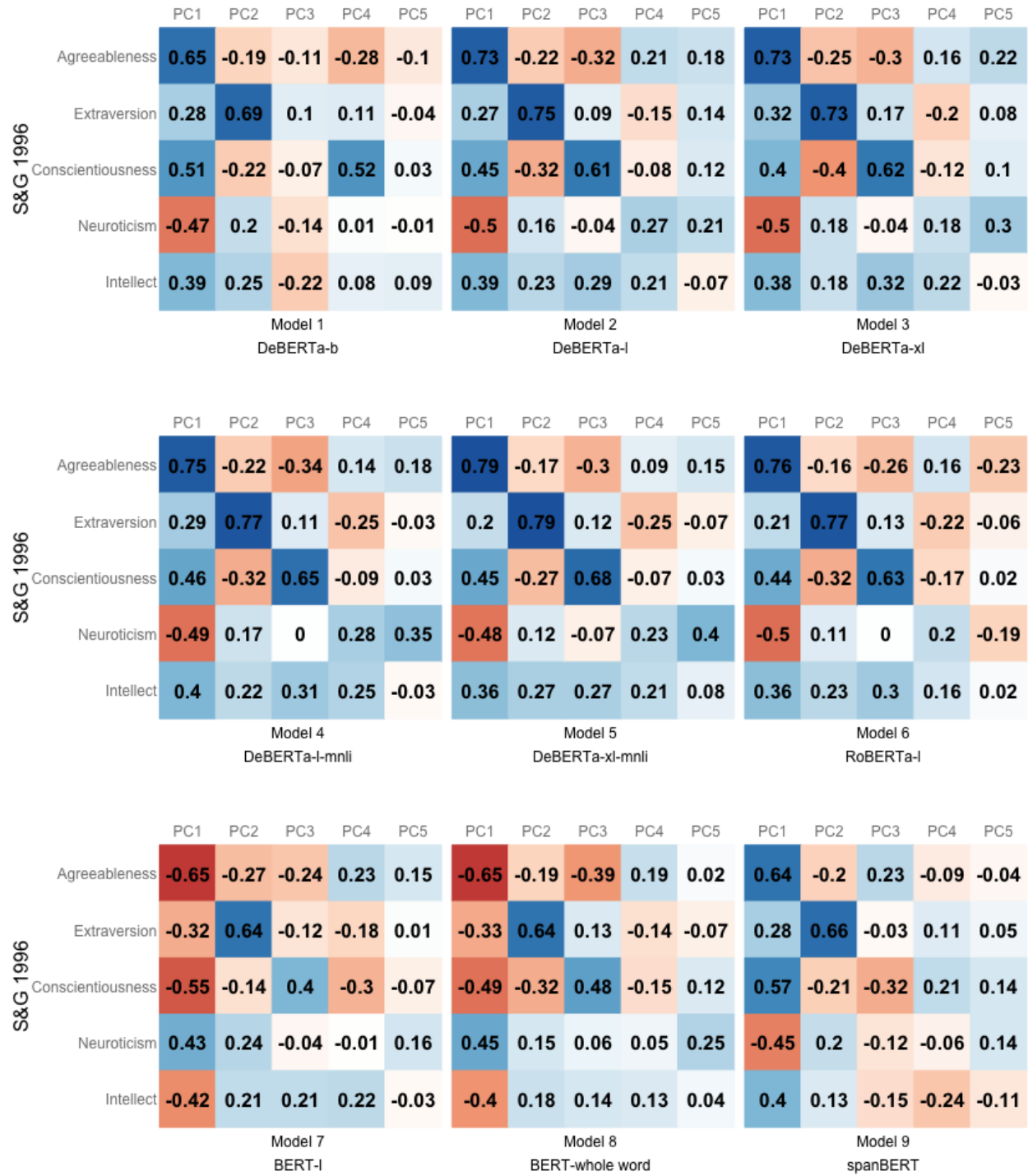

**Figure 11**

*Congruence coefficients between the second 9 (of 18) models and the S&G 1996 survey*

*components (Big Five).*

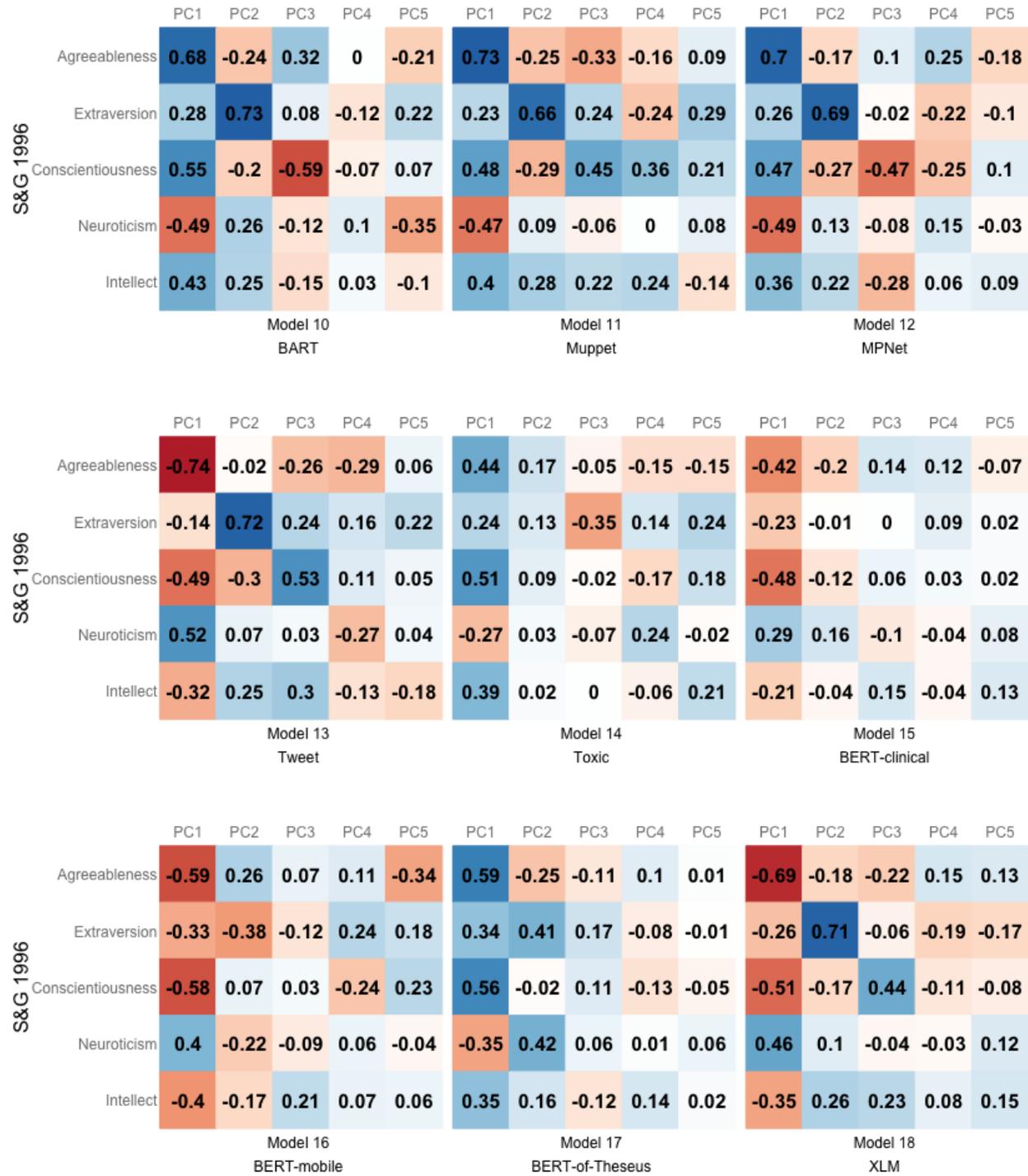

(.53 to .73 across all 3 components). Lower congruences were seen on the first 3 components with BERT-l (the predecessor to DeBERTa, RoBERTa, and others), DeBERTa-base (smaller in size), and other models that were optimized for different tasks and strategies (BERT whole word,

Muppet, MPNet, spanBERT, XLM). The two compressed models (BERT-of-Theseus and

BERT-mobile), Toxic, and BERT-clinical perform poorly. Note that none of the models recover

Neuroticism or Intellect as unique components; for all models, Neuroticism and Intellect are most congruent with the first component.

As described in more detail in the Supplementary Materials (Section 5.3), a post-hoc

analysis of the XLM model evaluated its potential for use in multilingual analyses of personality

structure. This was done by inputting the 435 terms in both English and Spanish simultaneously.[3]

Component analyses of the resulting data showed that the first component differentiated the

terms by language of origin, and the second component was Affiliation (see Supplemental Figure

29). The third component reflected features of Dynamism (energetic, imaginative, scrupulous

[negative], docile [negative]), and the fourth component conflated aspects of both Dynamism and

Order.

## Study 3 Discussion

Study 3 provided further insight on each of the three methodological decisions of Study

2. The set of personality descriptors seemed to have only a modest effect on the resulting

component structure. Both the smaller 83-term set and the larger 1,710-term set produced component structures that were highly similar to that produced by the 435-terms of S&G 1996, especially on the first 4 components. Even the much larger 18,000 term set was highly similar to the 435 on the first 3 components despite containing a high proportion of obscure descriptors.

---

[3] Google Translate was used to generate Spanish translations of the English terms, resulting in 371 unique
Spanish descriptors. A more robust approach would use an "emic"-generated list (Benet-Martínez, 2020).

Results from the study of different queries were similarly modest in terms of the effect on structure, despite the more subjective nature of generating queries. For 7 of the 8 queries, the first 3 components of Study 2 were robustly recovered. In this sub-study, there was some evidence of congruence with Study 2 on the fourth and fifth component, but follow-up analyses suggest that the orientation and content of these are inconsistent across queries. As in Study 2, the last two components also explain little variance.

The third sub-study showed that the choice of model is an influential decision. The largest and most recently developed models produced components that were most congruent with the Big Five. The specialized models generally performed less well, though these still provided some insight. For example, the performance of models trained on single sources of text varied depending on the source – the model trained on Twitter data did reasonably well and the model trained on electronic health records fared poorly. The Toxic model, trained to identify negative comments online, predictably recovered only a single factor. More compressed models failed to capture the full complexity of the more robust options. Though the multilingual model performed only moderately well in English by itself, results from the post-hoc analyses of English and Spanish together provide strong motivation for more rigorous multilingual evaluations, especially if these models improve enough to match the performance of DeBERTa. The ability to evaluate many – perhaps dozens – of languages simultaneously has the potential to rapidly accelerate the development of structural models of personality that generalize across cultures.

## General Discussion

The current work provides strong evidence of the utility of natural language processing models for personality structure research. This is based on the findings that the correlational structure of personality descriptors previously identified through the use of survey-based ratings is highly similar to the structure of these same terms in model-generated NLP data, and that the correlations in the NLP data had notably higher magnitudes. These findings were robust to sensitivity analyses for the queries and models in Study 3.

These results also point to important theoretical implications for personality measurement. For one, compelling external validation of the traditional psycholexical approach is provided by the finding that approximately 96% of all statistically significant associations in the data from S&G 1996 were directionally consistent with the NLP data in Study 2. Though the structure of ratings has historical precedent by 25 years, the NLP data are based on corpora of text that collectively capture a quantity of information many orders of magnitude greater than the survey ratings. These data confirm that self- and other-ratings of adjectives produce a structure that substantially reflects the semantic relations among personality descriptors, as hypothesized/presumed by many (Allport & Odbert, 1936; Ashton, Lee, & Goldberg, 2004; Cattell, 1943; Digman & Inouye, 1986; Goldberg, 1981, 1990, 1992; McCrae, 1990; McCrae & Costa, 1985; Norman, 1963; Thurstone, 1934; Tupes & Christal, 1961).

A related point is that the signal captured by surveys of adjective ratings is (a) mostly driven by semantic structure, and (b) relatively weak in comparison to methods that can describe this structure more directly. Though the string of responses provided by individual raters may represent unique *psychological* signatures of the targets being rated, most of the shared signal in large, noisy samples of raters seems to be semantic. This suggests that an acceptable strategy for developing nomothetic models of personality structure may be to focus on semantic structure explicitly, either independently or as a complement to the otherwise exclusive use of survey data.[4]

Additional findings relate to the lower-order dimensional structure evident in the language model data. As these were exploratory analyses with a new method, our interpretations should be seen as preliminary. Across all analyses, there was consistent evidence of two highly replicable components, and nearly as much evidence supporting the three-component solution. The exact content of these components remains an open question for subsequent work, but the preliminary evidence is aligned with the three factors of DeRaad et al. (Affiliation, Dynamism, and Order; De Raad et al., 2010) and Peabody (DeRaad & Peabody, 2002; Peabody, 1984). This means they are also overlapping with the two factor solutions of Saucier et al. (Social Self-Regulation and Dynamism; Saucier et al., 2014), and the meta-traits of Digman (Alpha and Beta; Digman, 1997) and DeYoung (Stability and Plasticity; DeYoung, 2006; DeYoung et al., 2002). See Section 2 of the Supplement for further discussion of the similarity of these constructs to many extant theories in psychology.

The lack of support for models with more than three components, especially the Big Five, is also notable. Though the five-component solution from S&G 1996 was exactly reproduced with the original data in Study 1, the follow-up analyses – bass-ackwards and evaluation of the effects of rotation – showed that the fourth and fifth components are much less clear than the first three. Further, these components do not replicate using the current method. In the language model data, the fourth and fifth components explain a trivial amount of variance (half as much as the third component), and their content is inconsistent and hard to interpret in Study 3. In our view, the substantial increase in signal provided by the language data makes the prior weak evidence for including these traits among the most parsimonious dimensions of personality look less compelling.[5]

---

[4] It is important to distinguish the interpretation of these results from the semantic similarity hypothesis advanced by D'Andrade (1974), Schweder (1975), and others. They argued that self- and other-ratings largely reflect the rater's schema for classifying meaning (based mainly on semantics, by default) rather than psychological features of the target (Block et al., 1979; D'Andrade, 1974; Romer & Revelle, 1984; Schweder, 1975). This idea is different from the current finding that the traditional psycholexical approach organizes traits similarly to the semantic structure of the descriptors (as intended, given the lexical hypothesis). A more relevant consideration is the apparent noise around the nomothetic structure in the S&G 1996 ratings data, relative to the stronger signal in the NLP data. Though it is possible that this noise stems from differences in cognitive schema across raters, it more likely results from the manifestation of individual differences -- that is, personality.

The lack of evidence from these data supporting the primacy of Neuroticism and Intellect/Openness does not diminish the significance of these traits. Both are widely (and reliably) assessed as part of existing assessment tools, and both are well-established as valid constructs. In fact, these circumstances point to the potential benefit of extracting many additional traits from psycholexical data beyond those that can be identified at the highest level of parsimony (e.g., Saucier & Iurino, 2019; Condon, 2018).

Among the recommendations for future work, further development of personality models along these lines is a top priority. Analytically, the empirical identification of lower-order traits need not be limited to the traditional psychometrics methods used here. Given the ability to evaluate much larger sets of personality descriptors and to access more informative data from DeBERTa and similar technologies, clustering approaches (Golino et al., 2020; Zadeh & Ben-

---

[5] Concerns about the primacy of Neuroticism and Intellect have previously been addressed by Goldberg. In his review of Big Five history (Goldberg, 1993), he claims to have wrestled for a decade with conflicting evidence supporting the extension of Peabody's three factor model towards Norman's five factor structure, going so far as describing his experience in this period as if "looking through a glass darkly" (p. 29). It is noteworthy that this occurred during an era when two theoretical 3 factor models were already widely used in research -- the NEO (Costa & McCrae, 1978) and P-E-N (Eysenck, 1970) models -- as both theories include a Neuroticism factor and one includes Openness to New Experiences (the NEO). In fact, Goldberg explains that the decision to expand the NEO to include Agreeableness and Conscientiousness was influenced by input from him and Digman during the period around meetings about the topic between 1981 and 1983 (p. 30, Goldberg, 1993).

David, 2012) may be more productive for identifying secondary traits than principal components or factor analyses.

Further evaluation of language modeling techniques is also needed – in multiple directions. One priority is to consider alternative methods for extracting information from existing models. We use a query that loads information onto a masked token, but not all language models allow for masking. It may be that alternative approaches to querying allow for even more robust identification of personality structure. These approaches may include methods that allow for evaluations of structure among personality items – phrases and sentences about behavior, affect, and cognition – in addition to trait descriptive adjectives.

Improved application of multilingual models would seem to be particularly useful for cross-cultural evaluations of personality structure. This is because they help to resolve the complication of needing to sample people in non-overlapping populations who speak only one or a few languages. The XLM model performed moderately well relative to other options in the English-only sample (though noticeably less well than DeBERTa), and it appeared to recover the familiar 2 component solution in a preliminary analysis of Spanish and English together. A third component similar to Order was evident, but less clear. More rigorous work is needed, as are evaluations of many additional languages, including low-resource languages and those that are no longer widely spoken.

Additional benefit is also likely from emergent technologies in language modeling. As the newer models performed better in Study 3.3 than those that are only a few years old, it is exciting to note that the state of the art in language modeling has advanced several orders of magnitude during the preparation of this work (from the ability to estimate 330 million parameters [RoBERTa; Liu et al., 2019] to 530 billion [Megatron; Narayanan et al., 2021]). Such rapid innovation suggests that it may be possible to extract increasingly fine representations of personality, including those that span modalities by connecting language to images or videos. For example, the CLIP language model (Mokady et al., 2021), trained on Instagram photos and their captions, is particularly promising considering how well the Twitter model performed in Study 3.

The current work has several limitations. Though we sought to evaluate the most impactful methodological decisions, several "forking paths" were not considered (Gelman & Loken, 2014). For example, it remains unclear how much the results were affected by the strategy of using queries with mask tokens. While the mask token approach seems particularly well-suited for the evaluation of structure based on (mostly single-word) trait descriptors, this strategy constrains the language model in a specific way that was both intentional (i.e., to match the survey-based ratings method) and limiting. This may affect the resultant structure. It is also important to note that this strategy was pursued based on prior work in personality structure research. More fundamentally novel approaches (such as those not using single-word descriptors) may produce different results, and these may be equally valid.

There is also wide variety in the available language models. We sampled only 18 models in Study 3.3 from the many thousands available on Hugging Face, and several of those tested were variations on the BERT/RoBERTa/DeBERTa design. Variability in performance across models is expected based on model size, language, training data, and other factors, and we were limited in our ability to test these thoroughly. It is also important to note that bias is not necessarily reduced in language models relative to the use of survey methods (Bordia et al., 2019; Nadeem et al., 2020), and this is a common criticism of the technologies that enable language models like those used here. Finally, the findings of this work are limited in capturing the full range of psychological individual differences in that some of these are perceived through modalities other than language (Bellugi et al., 1989; Paunonen et al., 1992).

*Conclusion*

The ability to evaluate personality structure with novel language modeling technologies is an exciting prospect, as it may open or reinvigorate lines of research that are currently limited by the constraints of surveys. In addition to the possibilities already mentioned, these may include the development of language models that are built for this purpose or for evaluating personality in highly specific contexts (i.e., historical contexts, hard-to-reach populations). We hope that the findings and resources of the current work will encourage these lines of research, and the broader pursuit of a more informed understanding of personality.